\pdfoutput=1
\documentclass[11pt]{article}

\usepackage[preprint]{acl}

\usepackage{times}
\usepackage{latexsym}
\usepackage[T1]{fontenc}
\usepackage[utf8]{inputenc}
\usepackage{microtype}
\usepackage{inconsolata}
\usepackage{graphicx}

\usepackage{amsmath}
\usepackage{booktabs}
\usepackage{enumitem}
\usepackage{tikz}
\usetikzlibrary{arrows.meta,shapes.geometric}
\definecolor{cblue}{HTML}{2A78D6}
\definecolor{corange}{HTML}{EB6834}
\definecolor{caqua}{HTML}{1BAF7A}
\usepackage[safe]{tipa}
\DeclareCaptionType{algorithm}[Algorithm][List of Algorithms]
\newcommand{\commrep}{Community Representative}
\newcommand{\phonebal}{Phoneme Balance}

\title{Structure Before Sampling: Community-Aware Core-Set Selection \\ for Data-Efficient Text-to-Speech}

\author{Mizbaul Haque Maruf and Muhammad Nur Yanhaona \\
  Department of Computer Science and Engineering \\
  Brac University, Dhaka, Bangladesh \\
  \texttt{mizbaul.haque.maruf@g.bracu.ac.bd}, \texttt{nur.yanhaona@bracu.ac.bd}}

\begin{document}
\maketitle

\begin{abstract}
Text-to-speech (TTS) corpora are costly to record, yet many utterances add little
new phonetic information. Core-set selection reduces this cost by choosing a
small training subset under a fixed audio-duration budget. We represent a corpus
as a phonotactic graph that links each utterance to its most phonemically similar
ones, and we first test whether this graph has structure. In Bangla and English
corpora, its clustering is 199 and 56 times that of a size-matched random graph,
and its modularity is more than twice that of a degree-preserving random graph.
We then propose \commrep{}, a selector that samples across graph communities and
spreads its choices within each one, starting from utterances rich in rare
phonemes. At every budget and in both languages, it covers more rare phoneme
bigrams than random and entropy-based selection, and this lead holds on held-out
utterances. TTS models trained on its 20\% core-sets have a significantly lower
character error rate (CER) than models trained on equal-duration random or
entropy-based subsets in both languages. When all models train for the same
number of epochs, the Bangla core-set model also outperforms full-corpus training
(3.93\% vs.\ 4.47\% CER) with $4.5\times$ less training time.
\end{abstract}

\section{Introduction}
\label{sec:intro}

The main cost of building a text-to-speech (TTS) system is its training corpus.
Much of this cost is spent on redundancy: a corpus large enough to cover the
phonotactics of a language contains many utterances that add little new
information for the model. \emph{Core-set selection} aims to find a small subset
that trains a model nearly as well as the full corpus.

\begin{figure*}[t]
\centering
\newcommand{\toygraph}{%
  \coordinate (a1) at (0.25,2.05); \coordinate (a2) at (0.80,2.30);
  \coordinate (a3) at (1.05,1.75); \coordinate (a4) at (0.55,1.45);
  \coordinate (a5) at (0.10,1.60);
  \coordinate (b1) at (1.75,2.25); \coordinate (b2) at (2.30,2.40);
  \coordinate (b3) at (2.50,1.90); \coordinate (b4) at (2.05,1.60);
  \coordinate (b5) at (1.60,1.85);
  \coordinate (c1) at (0.85,0.55); \coordinate (c2) at (1.20,0.95);
  \coordinate (c3) at (1.70,0.80); \coordinate (c4) at (1.55,0.30);
  \coordinate (c5) at (1.05,0.15);
  \foreach \u/\v in {a1/a2,a2/a3,a3/a4,a4/a5,a5/a1,a1/a4,a2/a4,
                     b1/b2,b2/b3,b3/b4,b4/b5,b5/b1,b1/b4,b2/b4,
                     c1/c2,c2/c3,c3/c4,c4/c5,c5/c1,c2/c4,c1/c3,
                     a3/c2,b4/c3,a3/b5}
    {\draw[black!25, line width=0.4pt] (\u) -- (\v);}%
}
\begin{tikzpicture}[
  font=\footnotesize,
  panel/.style={draw=black!12, fill=black!3, rounded corners=4pt},
  head/.style={anchor=north west, font=\footnotesize\bfseries, inner sep=0pt},
  card/.style={draw=black!25, fill=white, rounded corners=2pt, inner sep=2.5pt},
  chip/.style={draw=black!25, fill=white, rounded corners=1.5pt, inner sep=1.6pt},
  evalbox/.style={card, align=center, text width=22mm, font=\scriptsize},
  note/.style={font=\scriptsize, text=black!65},
  flow/.style={-{Stealth[length=2.2mm,width=2.4mm]}, line width=1.1pt, black!35},
  sarrow/.style={-{Stealth[length=1.4mm]}, line width=0.5pt, black!45},
  vtx/.style={circle, inner sep=0pt, minimum size=5pt, line width=0.5pt},
  seed/.style={star, star points=5, star point ratio=2.3, inner sep=0pt,
               minimum size=8.5pt, draw=white, line width=0.3pt},
]
\draw[panel] (0,0) rectangle (3.3,4.3);
\draw[panel] (3.65,0) rectangle (7.85,4.3);
\draw[panel] (8.2,0) rectangle (12.4,4.3);
\draw[panel] (12.75,0) rectangle (15.6,4.3);
\node[head] at (0.15,4.15) {(a) Featurize};
\node[head] at (3.8,4.15) {(b) Build \& test graph};
\node[head] at (8.35,4.15) {(c) Select core-set};
\node[head] at (12.9,4.15) {(d) Evaluate};
\foreach \x in {3.3,7.85,12.4} {\draw[flow] (\x+0.02,2.15) -- (\x+0.33,2.15);}
\node[card] (utt) at (1.65,3.35) {``the cat sat''};
\node (ipa) at (1.65,2.7) {\textipa{[D@ k\ae t s\ae t]}};
\node[chip] at (0.75,2.05) {\textipa{D@}};
\node[chip] at (1.35,2.05) {\textipa{k\ae}};
\node[chip] at (1.95,2.05) {\textipa{\ae t}};
\node[chip] at (2.55,2.05) {\textipa{s\ae}};
\draw[sarrow] (utt.south) -- (ipa.north);
\draw[sarrow] (ipa.south) -- (1.65,2.28);
\draw[sarrow] (1.65,1.83) -- (1.65,1.52);
\foreach \i/\h in {0/0.42,1/0.36,2/0.80,3/0.36,4/0.20,5/0.10}
  {\fill[black!40] (0.66+\i*0.36,0.55) rectangle ++(0.24,\h);}
\draw[black!30, line width=0.4pt] (0.55,0.55) -- (2.75,0.55);
\node[note] at (1.65,0.28) {TF-IDF vector};
\begin{scope}[shift={(4.45,1.2)}]
  \toygraph
  \foreach \n in {a1,a2,a3,a4,a5} {\node[vtx, fill=cblue, draw=white] at (\n) {};}
  \foreach \n in {b1,b2,b3,b4,b5} {\node[vtx, fill=corange, draw=white] at (\n) {};}
  \foreach \n in {c1,c2,c3,c4,c5} {\node[vtx, fill=caqua, draw=white] at (\n) {};}
\end{scope}
\node[font=\scriptsize] at (5.75,0.98) {$C$: \textbf{0.140} vs.\ 0.0007 (ER)};
\node[font=\scriptsize] at (5.75,0.62) {$Q$: \textbf{0.470} vs.\ 0.172 (config.)};
\node[note] at (5.75,0.25) {observed vs.\ null, Bangla};
\begin{scope}[shift={(9.0,1.2)}]
  \filldraw[fill=cblue!10, draw=cblue!45, line width=0.5pt]
    (0.55,1.83) ellipse [x radius=0.72, y radius=0.62];
  \filldraw[fill=corange!10, draw=corange!45, line width=0.5pt]
    (2.04,2.0) ellipse [x radius=0.66, y radius=0.58];
  \filldraw[fill=caqua!10, draw=caqua!45, line width=0.5pt]
    (1.27,0.55) ellipse [x radius=0.64, y radius=0.58];
  \toygraph
  \foreach \n in {a1,a4} {\node[vtx, fill=white, draw=cblue] at (\n) {};}
  \foreach \n in {b1,b2,b4} {\node[vtx, fill=white, draw=corange] at (\n) {};}
  \foreach \n in {c1,c2,c4} {\node[vtx, fill=white, draw=caqua] at (\n) {};}
  \foreach \n in {a5,a3} {\node[vtx, fill=cblue, draw=white] at (\n) {};}
  \node[vtx, fill=corange, draw=white] at (b5) {};
  \node[vtx, fill=caqua, draw=white] at (c3) {};
  \node[seed, fill=cblue] at (a2) {};
  \node[seed, fill=corange] at (b3) {};
  \node[seed, fill=caqua] at (c5) {};
  \node[font=\tiny, anchor=east, xshift=-3.5pt] at (a2) {1};
  \node[font=\tiny, anchor=east, xshift=-3pt] at (a5) {2};
  \node[font=\tiny, anchor=west, xshift=3pt] at (a3) {3};
\end{scope}
\node[seed, fill=black!55] at (8.5,0.95) {};
\node[note, anchor=west] at (8.62,0.95) {rare seed};
\node[vtx, fill=black!55, draw=white] at (10.0,0.95) {};
\node[note, anchor=west] at (10.1,0.95) {selected};
\node[vtx, fill=white, draw=black!55] at (11.3,0.95) {};
\node[note, anchor=west] at (11.4,0.95) {other};
\fill[cblue] (8.45,0.42) rectangle (9.84,0.64);
\fill[corange] (9.88,0.42) rectangle (11.12,0.64);
\fill[caqua] (11.16,0.42) rectangle (12.15,0.64);
\node[note] at (10.3,0.22) {budget $B_c \propto$ community duration};
\node[evalbox] (S) at (14.05,3.3) {Core-set $S$\\$\sum_{u\in S} d_u \geq B$};
\node[evalbox] (cov) at (14.05,2.05) {Rare-bigram\\coverage};
\node[evalbox] (tts) at (14.05,0.8) {Train TTS on $S$\\$\to$ ASR CER};
\draw[sarrow] (S.south) -- (cov.north);
\draw[sarrow] (S.east) -- ++(0.18,0) |- (tts.east);
\end{tikzpicture}
\caption{Overview. (a)~Utterances become TF-IDF vectors over IPA phonemes and
within-word bigrams (\S\ref{ssec:features}). (b)~The $k$-NN utterance graph is
tested against Erd\H{o}s--R\'enyi (ER) and configuration-model nulls before
selection relies on it (\S\ref{sec:analysis}); colours mark Louvain communities.
(c)~\commrep{} splits the duration budget across communities and selects within
each in farthest-point order from the most rare-phoneme-rich utterance (star;
\S\ref{sec:method}). (d)~Core-sets are scored by rare-bigram coverage and
downstream TTS intelligibility (\S\ref{sec:setup}--\ref{sec:results}).}
\label{fig:pipeline}
\end{figure*}
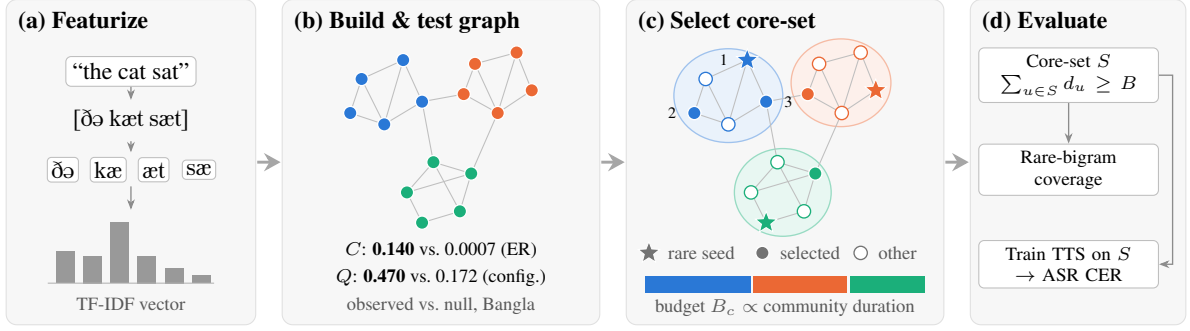

Existing data selection methods treat a corpus as a set of independent items.
They either choose sentences with balanced phoneme statistics
\citep{baspro,eusipco2022} or spread the selection across a feature space
\citep{sener2018}. This view ignores a well-documented
property of language: linguistic units form networks with strong community
structure and small-world geometry \citep{ferrer2001,vitevitch2008,siew2013}. If
a corpus inherits this organization, a selector that ignores it discards useful
information. If it does not, a graph-based selector is only an elaborate form of
random sampling.

We therefore treat the question empirically and answer it before building on it
(Figure~\ref{fig:pipeline}). We represent each utterance by its phonotactic
features, connect it to its nearest neighbours, and test the resulting graph
against matched null models. The graph is strongly clustered and modular, so we
derive a selector, \commrep{}, that stratifies the pool by graph community and,
within each community, spends the duration budget on well-spread utterances,
starting from the one richest in rare phonemes. We evaluate the selector
intrinsically, by the coverage of rare phoneme bigrams, and extrinsically, by the
intelligibility of TTS models trained on the selected subsets.

A single corpus cannot show whether this structure is a property of the language
or of one dataset. We therefore run the full study on two typologically distant
languages, Bangla and English \citep{ljspeech2017}, and share every
hyperparameter of the selection pipeline between them, so that differences
between the two reflect the language and the corpus rather than tuning. Our
contributions are as follows:

\begin{itemize}[leftmargin=*,nosep,topsep=3pt]
\item \textbf{Structure, tested rather than assumed.} We build a phonotactic
$k$-nearest-neighbour graph over utterances and show that its clustering is
$199\times$ the Erd\H{o}s--R\'enyi value in Bangla and $56\times$ in English, and
that its modularity exceeds a degree-preserving configuration model in both
(\S\ref{sec:analysis}).
\item \textbf{A structure-aware selector.} \commrep{}
(\S\ref{sec:method}) covers more rare phonotactic patterns than random and
entropy-based selection at every budget, in both languages, on held-out data, and
on a third corpus with multiple speakers
(\S\ref{ssec:coverage}--\ref{ssec:heldout}).
\item \textbf{Controlled downstream evidence in two languages.} At an identical
duration budget, TTS models trained on our core-sets are significantly more
intelligible than models trained on a random subset or on a \phonebal{} subset,
in both Bangla and English. In Bangla, the core-set model also outperforms
full-corpus training with $4.5\times$ less training time
(\S\ref{ssec:tts_bn}--\ref{ssec:tts_en}).
\end{itemize}

\section{Related Work}
\label{sec:related}

\paragraph{Data selection for TTS.}
Data selection for TTS has mainly been studied as script selection for corpus
collection. These methods choose candidate sentences so that the selected set
has balanced phonetic content, for example with a genetic algorithm
\citep{baspro} or with dedicated procedures for selecting phonetically balanced
sentences \citep{eusipco2022}. Our \phonebal{} baseline (\S\ref{ssec:baselines})
follows the same principle. These methods score sentences by their phoneme
statistics, but they do not model how utterances relate to one another, which is
the information our graph captures.

\paragraph{Core-set and budgeted selection in machine learning.}
In computer vision, \citet{sener2018} formulate core-set selection as a
$k$-center problem over embeddings. We reuse their greedy $k$-center solver but
apply it \emph{within} communities rather than over the whole pool, so that the
geometric criterion works inside strata defined by the graph. More broadly,
budgeted subset selection is commonly solved greedily, with approximation
guarantees for submodular objectives \citep{nemhauser1978} and extensions to
node selection in networks \citep{kempe2003,leskovec2007}. These methods take
the network as given, whereas we first test whether the graph has structure
worth exploiting.

\paragraph{Data pruning and deduplication.}
Recent work reduces training cost by pruning large datasets. \citet{sorscher2022}
show that a good pruning metric can make error fall faster than the usual power
law in dataset size, and \citet{zheng2023ccs} retain accuracy at high pruning
rates by jointly considering data coverage and example importance.
\citet{maharana2024d2} also represent a dataset as a graph, and they select a
core-set by message passing that balances diversity and difficulty.
Deduplication removes near-duplicates from language-model training data
\citep{lee2022dedup} and semantic duplicates from web-scale image--text data
\citep{abbas2023semdedup}. For speech, dynamic data pruning matches full-data ASR
performance while training on 70\% of the data \citep{xiao2024ddp}. These
methods score examples with trained models, training dynamics, or duplicate
detection. Our selector needs no model trained on the data and relies on the
phonotactic content of the transcripts, and, unlike \citet{maharana2024d2}, we
test whether the graph has non-random structure before selecting on it.

\paragraph{Network structure of language.}
Word co-occurrence networks are small worlds \citep{ferrer2001} in the sense of
\citet{watts1998}. Phonological networks of the lexicon show community structure
\citep{siew2013} and have been used to study word learning and lexical retrieval
\citep{vitevitch2008}. We ask whether a TTS corpus, viewed as a graph of
utterances, shares this organization.

\paragraph{Community detection and null models.}
We detect communities with Louvain modularity optimization \citep{blondel2008}
and assess structure against Erd\H{o}s--R\'enyi and configuration-model nulls
\citep{barabasi2016}. Both are standard tools in network science. Our
contribution is to apply them to TTS corpus design, to test the graph before
relying on it, and to check that both the test and the downstream result hold
across languages.

\section{Problem Formulation and Graphs}
\label{sec:formulation}

\subsection{Duration-Budgeted Core-Set Selection}
\label{ssec:problem}

Let $U$ be a pool of transcribed utterances, where utterance $u$ has audio
duration $d_u$. Given a budget fraction $\beta$, core-set selection seeks a
subset $S \subseteq U$ with
\begin{equation}
\textstyle\sum_{u \in S} d_u \;\geq\; B = \beta \sum_{u \in U} d_u
\label{eq:budget}
\end{equation}
such that a TTS model trained on $S$ performs nearly as well as one trained on
$U$. We express budgets in \emph{audio duration} rather than utterance count,
because training cost scales with duration and a count budget would reward
methods that prefer short utterances. We score subsets intrinsically by their
coverage of rare phoneme bigrams
(\S\ref{ssec:intrinsic_eval}) and extrinsically by the intelligibility of TTS
models trained on them (\S\ref{ssec:tts_train}--\ref{ssec:tts_eval}).

\subsection{Corpora}
\label{ssec:corpora}

For Bangla, we use the verified sentences of Common Voice Scripted Speech 26.0
-- Bengali \citep{ardila2020}, recorded by a single in-house speaker (40{,}422
utterances, 48.75 hours). For English, we use LJSpeech-1.1 \citep{ljspeech2017}
(13{,}100 utterances, 23.92 hours, a single human reader). Both corpora are
sampled at 22.05\,kHz and contain one speaker each, so no objective includes a
speaker-coverage term. In each corpus, a seeded 10\% split is held out from
feature extraction, graph construction, the rarity signal, and all selectors, and
it supplies the evaluation items. Table~\ref{tab:corpus} describes both corpora and their
graphs. A third, multi-speaker corpus is used only to test transfer
(\S\ref{ssec:heldout}).

\paragraph{Shared configuration.} All numeric hyperparameters are shared between
the two languages: $k=15$, budgets of 10/20/40\% of pool duration, five seeds, a
10\% held-out fraction, phoneme unigram and bigram features, and a
bottom-quartile definition of \emph{rare}. None of them was tuned per language.

\begin{table}[t]
\centering
\caption{Corpus and graph descriptors. Every hyperparameter behind these numbers
is shared across the two languages.}
\label{tab:corpus}
\small
\setlength{\tabcolsep}{3pt}
\begin{tabular}{lrr}
\hline
quantity & Bangla & English \\
\hline
utterances / hours & 40{,}422 / 48.75 & 13{,}100 / 23.92 \\
pool / held-out & 36{,}389 / 4{,}033 & 11{,}769 / 1{,}331 \\
IPA phonemes & 37 & 41 \\
within-word bigrams & 775 & 1{,}073 \\
rare bigrams (bottom qu.) & 194 & 278 \\
rare share of tokens (\%) & 0.099 & 0.351 \\
TF-IDF dimension & 812 & 1{,}114 \\
$G_u$ nodes & 36{,}388 & 11{,}769 \\
$G_u$ edges & 462{,}066 & 150{,}416 \\
$G_u$ $\langle k\rangle$ / $\ln N$ & 25.40 / 10.50 & 25.56 / 9.37 \\
$G_u$ Louvain communities & 27 & 17 \\
$G_p$ nodes / edges & 37 / 465 & 41 / 653 \\
\hline
\end{tabular}

\end{table}

\subsection{Phonotactic Features}
\label{ssec:features}

We convert text to IPA. For Bangla, we use \texttt{epitran} \citep{epitran2018}
with the \texttt{ben-Beng-east} model after Unicode normalization. For English,
we use a CMUdict \citep{cmudict} lookup with a neural grapheme-to-phoneme model
for out-of-vocabulary words, and we map ARPAbet symbols to IPA. We remove English
stress digits with one exception: the unstressed \texttt{AH0} and \texttt{ER0}
are kept distinct from the stressed \texttt{AH1} and \texttt{ER1}, so that schwa
and r-coloured schwa remain separate from the wedge and the stressed r-coloured
vowel. CMUdict marks reduced vowels only through the stress digit. Removing all
digits would therefore merge schwa, the most frequent English phoneme (76{,}134
tokens in the pool), into the wedge and distort every distributional metric we
report.

Each utterance is represented as a TF-IDF vector over phoneme unigrams and
bigrams, $L_2$-normalized so that cosine similarity is a dot product. Bigrams
are counted \emph{within words only}. A word-final $/k/$ followed by a
word-initial $/t/$ is not a cluster in either language, and counting it would
inflate every coverage number. Only 775 phoneme pairs are attested in Bangla and
1{,}073 in English, which reflects the phonotactics of each language. No acoustic
embedding is used; audio contributes only its duration.

\subsection{Graph Construction}
\label{ssec:graphs}

We build two graphs per language. The \emph{utterance graph} $G_u$ has as nodes
the pool utterances that contain at least one phoneme, and each node is joined
to its $k=15$ exact nearest neighbours by cosine similarity. The \emph{phoneme
graph} $G_p$ has phonemes as nodes, which are joined when they are adjacent
within a word and weighted by co-occurrence. We use $G_p$ only for analysis
(\S\ref{ssec:gp}). For selection, we score each utterance by the rare phonemes it
contains:
\begin{equation}
r_u = \textstyle\sum_{p \in u} 1/\log(2+f_p),
\label{eq:rarity}
\end{equation}
where $f_p$ is the frequency of phoneme $p$ in the pool and the sum runs over the
distinct phonemes of $u$. The score depends only on phoneme frequencies, and it
is larger for utterances that contain rarer phonemes.

We set $k=15$ from theory rather than by tuning, since tuning $k$ on the
evaluation metric would be circular. A random graph is almost surely connected
once $\langle k\rangle > \ln N$ \citep{barabasi2016}. After symmetrization,
$\langle k\rangle = 25.40$ against $\ln N = 10.50$ in Bangla and $25.56$ against
$9.37$ in English. Both graphs are therefore in the connected regime, with a
giant component that covers every node. No community is isolated from the rest
of the pool, and every utterance is reachable by the selector.

\begin{table*}[t]
\centering
\caption{$G_u$ against matched nulls in both languages (ER $=$
Erd\H{o}s--R\'enyi, config.\ $=$ degree-preserving configuration model),
averaged over 5 realizations. Clustering $C$ and modularity $Q$ exceed both
nulls in both languages.}
\label{tab:graph}
\small
\setlength{\tabcolsep}{4pt}
\begin{tabular}{lcccccc}
\hline
& \multicolumn{3}{c}{Bangla} & \multicolumn{3}{c}{English} \\
\cmidrule(lr){2-4}\cmidrule(lr){5-7}
& $G_u$ & ER & config. & $G_u$ & ER & config. \\
\hline
$N$ & 36{,}388 & matched & matched & 11{,}769 & matched & matched \\
$L$ & 462{,}066 & matched & matched & 150{,}416 & matched & matched \\
$\langle k\rangle$ & 25.4 & 25.4 & 25.4 & 25.6 & 25.6 & 25.5 \\
$C$ & \textbf{0.1401} & 0.0007 & 0.0018 & \textbf{0.1211} & 0.0022 & 0.0056 \\
$Q$ & \textbf{0.470} & 0.156 & 0.172 & \textbf{0.404} & 0.155 & 0.177 \\
$\langle k^2\rangle/\langle k\rangle^2$ & 1.65 & 1.04 & 1.64 & 1.65 & 1.04 & 1.63 \\
$k_{\max}$ & 627 & 51 & 618 & 461 & 48 & 447 \\
$\langle d\rangle$ & 3.72 & 3.61 & 3.40 & 3.23 & 3.19 & 3.05 \\
\hline
\end{tabular}

\end{table*}

\begin{figure*}[t]
\centering
\includegraphics[width=\textwidth]{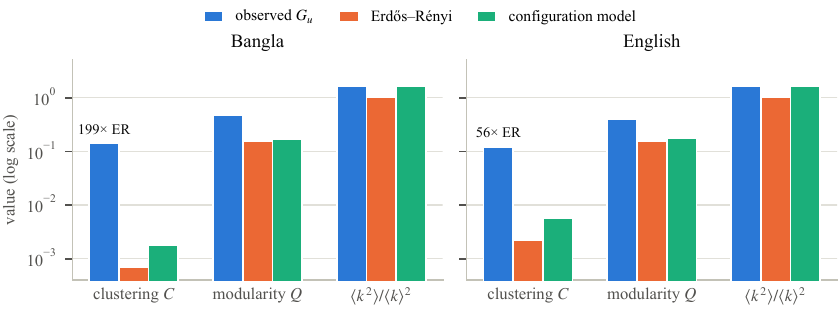}
\caption{$G_u$ versus matched nulls (log scale), one panel per language.
Clustering and modularity both exceed the degree-preserving configuration model
in Bangla and in English.}
\label{fig:nulls}
\end{figure*}

\section{Structural Analysis: Is the Structure Real?}
\label{sec:analysis}

Before building a selector on $G_u$, we test whether it differs from a random
graph. If it did not, Louvain stratification would amount to random sampling, and
\phonebal{} would be the better choice.

\begin{figure*}[t]
\centering
\includegraphics[width=\textwidth]{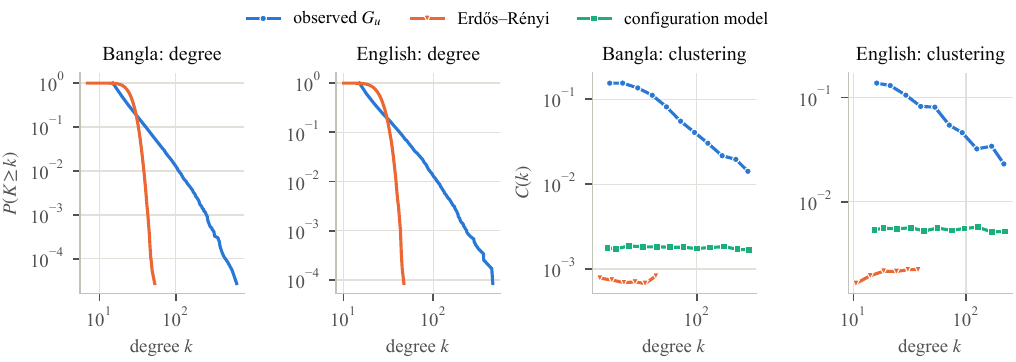}
\caption{Degree distribution and clustering spectrum of $G_u$ against one
realization (seed 13) of each null model. $P(K \geq k)$ is the complementary
cumulative degree distribution; $C(k)$ is the mean local clustering coefficient
of utterances in logarithmic degree bins. The configuration model shares $G_u$'s
degree sequence, so it appears only in the clustering panels.}
\label{fig:degree}
\end{figure*}

\subsection{Null Models}
\label{ssec:nulls}

We use two matched null models, each averaged over five realizations: an
Erd\H{o}s--R\'enyi graph $G(N,L)$ with the same numbers of nodes and edges, and a
\emph{configuration model} that preserves the exact degree sequence. The second
is the stronger test, because structure that survives it cannot be explained by
degree heterogeneity alone. We compare clustering $C$, Louvain modularity $Q$,
degree heterogeneity $\langle k^2\rangle/\langle k\rangle^2$, maximum degree
$k_{\max}$, and mean path length $\langle d\rangle$.

\subsection{The Utterance Graph Is Clustered, Modular and Small-World}
\label{ssec:gu}

Table~\ref{tab:graph} and Figure~\ref{fig:nulls} report the results. The same
three findings hold in both languages.

\paragraph{Clustering.} $C$ is far above the random value: $0.140$ against
$0.0007$ in Bangla ($199\times$) and $0.121$ against $0.0022$ in English
($56\times$). It remains $79\times$ and $22\times$ higher than in the
degree-preserving null. Both graphs are therefore strongly triadic, which makes
the coverage of local neighbourhoods a meaningful objective.

\paragraph{Modularity.} $Q$ exceeds the degree-preserving null in both
languages: $0.470$ against $0.172$ in Bangla and $0.404$ against $0.177$ in
English. The 27 and 17 detected communities therefore reflect genuine grouping in
phonotactic space rather than a degree artifact. This result motivates
stratifying the selection by community, and it mirrors the finding of
\citet{siew2013} for phonological networks.

\paragraph{Small-world geometry.} Mean path length is close to the random value
in both languages ($3.72$ against $3.61$ in Bangla; $3.23$ against $3.19$ in
English), while clustering is one to two orders of magnitude higher. This is the
small-world signature of \citet{watts1998}, and it places both corpora in the
same structural family as the word networks of \citet{ferrer2001}.

\paragraph{Hubs and local clustering.} Figure~\ref{fig:degree} examines these
averages node by node. The degree distribution of $G_u$ has a heavy tail that
the matched random graph lacks. In both languages, 1.3\% of utterances have
degree 100 or more, while the Erd\H{o}s--R\'enyi realization has none, and the
largest degree is 627 in Bangla and 461 in English, against 53 and 47 in this
realization (51 and 48 on average over the five realizations in
Table~\ref{tab:graph}). Clustering
decreases with degree, from 0.154 at $k \approx 16$ to 0.014 at $k \approx 291$
in Bangla and from 0.137 to 0.023 in English, so hubs are the least clustered
utterances. The configuration model stays flat at every degree, and $G_u$ remains
at least $8\times$ above it in Bangla and $4\times$ in English across the whole
degree range. The triadic structure is therefore not driven by a few hubs but
holds throughout the graph.

English is weaker than Bangla on every structural measure: its graph is less
clustered and less modular, and its corpus is about half the size in hours. The
small-world signature is nonetheless clear, and \S\ref{ssec:coverage} shows that
the selection advantage it predicts also holds in English.

\subsection{The Phoneme Graph Is Disassortative}
\label{ssec:gp}

$G_p$ requires a different test. It is small and nearly complete in both
languages (37 nodes at density $0.70$ in Bangla and 41 nodes at density $0.80$ in
English), so a comparison with a Poisson degree distribution is uninformative,
and no scale-free claim can be made on so few nodes. Its structure lies instead
in \emph{which} phonemes connect to which. Degree assortativity is $-0.339$ in
Bangla and $-0.122$ in English, against $-0.048$ and $-0.049$ for matched random
graphs. This disassortativity indicates a hub-and-spoke organization, in which
well-connected phonemes attach preferentially to poorly connected ones rather
than to each other. We report it as a descriptive property of the phoneme
inventory; the selector does not use the edges of $G_p$.

\begin{figure*}[t]
\centering
\includegraphics[width=0.86\textwidth]{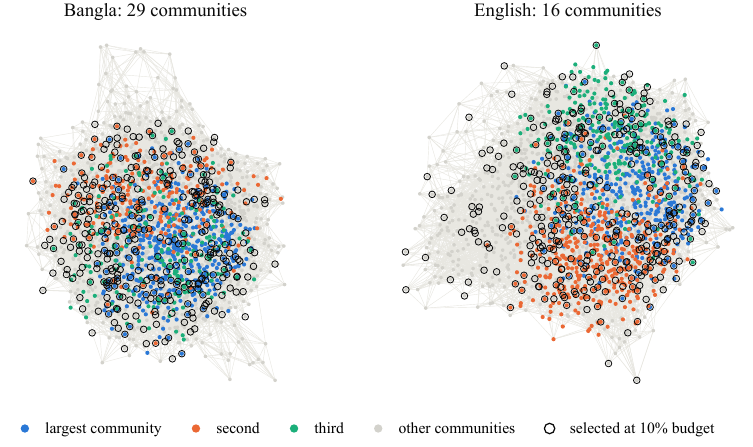}
\caption{Communities of $G_u$ and the utterances \commrep{} selects at a 10\%
budget. Communities are the Louvain partition the selector uses on the full
similarity-weighted graph. Because an induced subgraph of $G_u$ on a sample has
almost no edges, each panel lays out a $k$-NN graph ($k{=}8$) rebuilt on a
2{,}500-utterance sample stratified by community; the three largest communities
are coloured and the rest grey. Rings mark selected utterances.}
\label{fig:communities}
\end{figure*}

\section{Community-Aware Core-Set Selection}
\label{sec:method}

\commrep{} turns the findings of \S\ref{sec:analysis} into design choices. It and
both baselines receive the same input and fill the same duration budget
(Eq.~\ref{eq:budget}), so all selectors are compared at equal audio duration
rather than at equal utterance count.

\paragraph{Stratify by community.} Because $G_u$ is strongly modular
(\S\ref{ssec:gu}), we partition it with weighted Louvain \citep{blondel2008},
using cosine similarities as edge weights, and treat each community as a stratum.
Sampling \emph{across} communities spreads the budget over all regions of the
graph.

\paragraph{Spread within each community.} Within each community, members are
ordered by farthest-point sampling under cosine distance, which is the greedy
$k$-center solver \citep{sener2018}. Farthest-point sampling \emph{within} a
community provides spread.

\paragraph{Seed toward the tail.} The farthest-point order of each community
starts from its member with the highest rarity score $r_u$ (Eq.~\ref{eq:rarity}).
Phonotactic frequencies are highly skewed: the rare bigrams defined in
\S\ref{ssec:intrinsic_eval} make up only 0.099\% of bigram tokens in Bangla and
0.351\% in English, so random draws seldom reach them. Starting each stratum from
its rarest utterance places this tail early in the selection.

\paragraph{Allot the budget.} Each community receives a share of the budget
proportional to its total duration, and utterances are drawn round-robin across
communities. Algorithm~\ref{alg:commrep} gives the full procedure, and
Figure~\ref{fig:communities} shows its output. Louvain on the
similarity-weighted graph finds 29 communities in Bangla and 16 in English (the
unweighted partition in \S\ref{ssec:gu} has 27 and 17). At a 10\% budget, every
community contributes selected utterances: between 5.7\% and 15.1\% of its
members in Bangla and between 11.1\% and 15.9\% in English. The selections are
spread across the whole layout, including its periphery.

\begin{algorithm}[t]
\hrule height 0.8pt\vspace{2pt}
\caption{\commrep{} selection. A community $c$ is \emph{open} while
$\pi_c \neq \emptyset$ and $\mathrm{used}_c < B_c$.}
\label{alg:commrep}
\hrule\vspace{3pt}
{\small\setlength{\baselineskip}{14pt}
\begin{tabbing}
10~\=xx\=xx\=\kill
\textbf{input:} $G_u$, durations $d$, rarity $r$, budget $B$ (s),
  $B \leq \textstyle\sum_{v} d_v$\\
\textbf{output:} $S$ with $\textstyle B \leq \sum_{u \in S} d_u < B + \max_u d_u$\\
1\> $P \leftarrow \textsc{Louvain}(G_u)$, weighted by cosine sim.\\
2\> \textbf{for} each community $c \in P$ \textbf{do}\\
3\> \> $s_c \leftarrow \arg\max_{u \in c} r_u$ \hfill \textit{// rare seed}\\
4\> \> $\pi_c \leftarrow$ farthest-point order of $c$ from $s_c$\\
5\> \> $B_c \leftarrow B \cdot (\sum_{u \in c} d_u) / (\sum_{v} d_v)$\\
6\> $S \leftarrow \emptyset$; \; $\mathrm{used}_c \leftarrow 0$ for all $c$\\
7\> \textbf{while} $\sum_{u \in S} d_u < B$ and some $c$ is open \textbf{do}\\
8\> \> \textbf{for} each open $c$, while $\sum_{u \in S} d_u < B$ \textbf{do}\\
9\> \> \> pop $u$ from $\pi_c$ into $S$; \; $\mathrm{used}_c \leftarrow \mathrm{used}_c + d_u$\\
10\> \textbf{return} $S$
\end{tabbing}}
\vspace{-4pt}\hrule height 0.8pt
\end{algorithm}

\begin{table*}[t]
\centering
\caption{Downstream training arms: utterances / hours per language, and training
schedules. The three 20\% subsets have identical duration.}
\label{tab:arms}
\small
\setlength{\tabcolsep}{5pt}
\begin{tabular}{lcccc}
\hline
 & Community Representative & Phoneme Balance & Random & Full data \\
 & (20\%) & (20\%) & (20\%) & (100\%) \\
\hline
Bangla utt.\ / hours & 9{,}183 / 8.78 & 8{,}872 / 8.78 & 7{,}299 / 8.78 & 40{,}422 / 48.75 \\
English utt.\ / hours & 3{,}017 / 4.30 & 2{,}825 / 4.30 & 2{,}342 / 4.30 & 11{,}769 / 21.48 \\
\hline
Bangla schedule & \multicolumn{4}{c}{1500 epochs; batch 128 (Random: 98)} \\
English schedule & \multicolumn{4}{c}{30{,}000 gradient steps; batch 256} \\
\hline
\end{tabular}

\end{table*}

\begin{table*}[t]
\centering
\caption{Rare-bigram coverage (\%) by method, budget and language, mean $\pm$
95\% CI over 5 seeds. Entries without an interval have zero seed variance
(\S\ref{ssec:baselines}). Best per column in bold.}
\label{tab:intrinsic}
\small
\setlength{\tabcolsep}{4pt}
\begin{tabular}{lcccccc}
\hline
& \multicolumn{3}{c}{Bangla} & \multicolumn{3}{c}{English} \\
\cmidrule(lr){2-4}\cmidrule(lr){5-7}
method & 10\% & 20\% & 40\% & 10\% & 20\% & 40\% \\
\hline
\textbf{Community Representative} (ours) & \textbf{89.5\,$\pm$\,0.6} & \textbf{96.0\,$\pm$\,0.8} & \textbf{98.5\,$\pm$\,0.5} & \textbf{73.8\,$\pm$\,1.2} & \textbf{86.0\,$\pm$\,1.1} & \textbf{95.7\,$\pm$\,0.8} \\
Phoneme Balance & 73.2 & 82.5 & 93.8 & 55.0 & 71.9 & 86.0 \\
Random & 49.9\,$\pm$\,4.0 & 69.8\,$\pm$\,1.8 & 83.7\,$\pm$\,2.1 & 46.3\,$\pm$\,2.0 & 64.8\,$\pm$\,3.7 & 80.6\,$\pm$\,3.9 \\
\hline
\end{tabular}

\end{table*}

\begin{figure*}[t]
\centering
\includegraphics[width=\textwidth]{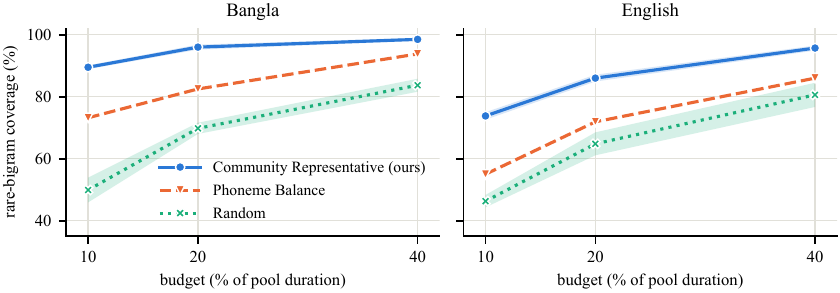}
\caption{Rare-bigram coverage vs.\ duration budget in both languages; bands are
95\% CIs over 5 seeds. The ranking of methods is the same in both panels.}
\label{fig:rare}
\end{figure*}

\section{Experimental Setup}
\label{sec:setup}

\subsection{Baselines}
\label{ssec:baselines}

Both baselines fill the same duration budget as \commrep{}.

\paragraph{\phonebal{}.} This baseline selects for phonetic balance. Let
$\mathbf{c}_S$ be the phoneme unigram counts of a selection $S$, and let
$H(\mathbf{c})$ be the entropy of the phoneme distribution with counts
$\mathbf{c}$. Starting from one utterance drawn with the run's seed, \phonebal{}
repeatedly adds
\begin{equation}
u^{*} = \textstyle\arg\max_{u \notin S} H(\mathbf{c}_S + \mathbf{c}_u)
\label{eq:phonebal}
\end{equation}
until the budget is filled \citep{seki2024}. It therefore favours subsets whose
phoneme distribution is as uniform as possible. Only its first pick depends on
the seed, and its coverage does not vary across our five seeds.

\paragraph{Random.} Random samples utterances uniformly without replacement,
with a fixed seed, until the budget is filled.

\subsection{Intrinsic Evaluation}
\label{ssec:intrinsic_eval}

We report the coverage of phoneme \emph{bigram} types. Unigram coverage is $1.0$
for every method at every budget in both languages and is therefore
uninformative. The main metric is the coverage of \emph{rare} bigrams, defined as
the bottom quartile by frequency: 194 types in Bangla, which account for only
0.099\% of bigram tokens, and 278 types in English, which account for 0.351\%.
Random sampling tends to miss exactly these types. Each method and budget is run
with 5 seeds, and we report the mean and 95\% confidence interval (CI).

We add three further measurements. \emph{Held-out coverage} controls for
circularity: a method could win on pool coverage by exploiting the statistics it
was built from, but it cannot do so on the bigram types of held-out utterances.
This control matters because \commrep{} spreads its selection over phoneme
unigram and bigram features, so coverage on the pool is closely related to what it
optimizes.
\emph{Multi-speaker transfer} repeats the intrinsic evaluation on OpenSLR SLR37
(1{,}891 utterances, 2.94 hours, six Bangla speakers), because both main corpora
are single-speaker. \emph{Graph coverage} measures selection on $G_u$ itself: the
share of pool utterances that are selected or are among the $k{=}15$ nearest
neighbours of a selected utterance, computed at every 2\% of budget. Because
\commrep{} selects on this graph and the baselines do not, graph coverage partly
favours it by construction, so we treat it as a description of the selections
rather than as independent evidence.

\subsection{Downstream TTS Training}
\label{ssec:tts_train}

In each language, we train a lightweight TTS model, MB-iSTFT-VITS
\citep{mbistft2023,vits2021}, on the \commrep{} core-set at a 20\% duration
budget and on three controls: the full training data, a random subset, and a
\phonebal{} subset, with both subsets at the same duration as the core-set. The
random arm separates the effect of the selection method from the effect of the
budget. The \phonebal{} arm is the stricter control, since it fixes the budget and
changes only the selection criterion. Table~\ref{tab:arms} lists the arms. Equal
duration does not imply an equal number of utterances: \commrep{} prefers shorter
utterances and therefore selects about a quarter more of them than Random in both
languages. Duration is still the budget that reflects cost (\S\ref{ssec:problem}).
All arms use a learning rate of $2\times10^{-4}$, seed 1234, and fp16 training.

\paragraph{Bangla: equal epochs.} The Bangla arms are matched on epochs, 1500
each. Since an epoch is one pass over an arm's own audio, equal epochs make total
compute proportional to the duration budget. This is the source of the cost
saving: the core-set run takes 109{,}400 steps in 11.3\,h and the random run
110{,}900 steps in 10.7\,h, against 454{,}400 steps in 50.4\,h for the full
corpus, a $4.5\times$ reduction in wall-clock time. The \phonebal{} arm uses the
same schedule and batch size as the core-set arm, so the two are compared at
equal budget \emph{and} equal compute. One hyperparameter differs across the
Bangla arms: the random arm uses a batch size of 98 instead of 128. We chose this
value so that the core-set and random arms end within 1.4\% of each other in
gradient steps, even though they contain different numbers of utterances.

\paragraph{English: equal updates.} All four English arms use a batch size of 256
and are trained for a fixed 30{,}000 gradient steps, set in advance. Every arm
therefore receives the same number of updates at the same batch size, so the
English comparison is one of equal compute rather than of equal passes over data
of different sizes. As a result, the two languages answer related but different
questions, which we discuss in \S\ref{ssec:epochs}. In both languages, training
loss is computed on each arm's own training data and is not comparable across
arms, so we compare the arms only through held-out intelligibility
(\S\ref{ssec:tts_eval}).

\begin{figure*}[t]
\centering
\includegraphics[width=\textwidth]{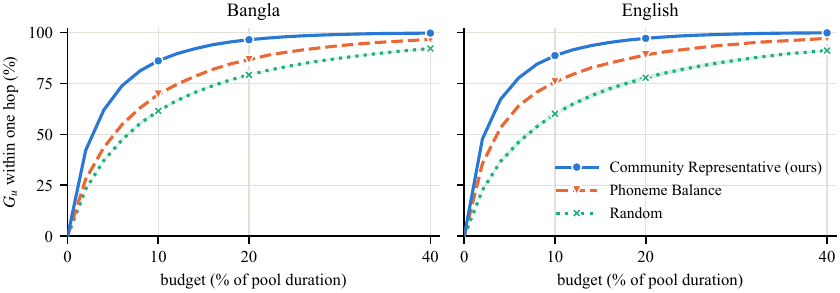}
\caption{Graph coverage against the duration budget: the share of pool
utterances that are selected or among the 15 nearest neighbours of a selected
utterance. \commrep{} is re-run at every 2\% budget; bands are 95\% CIs over 5
seeds.}
\label{fig:graphcov}
\end{figure*}

\subsection{Downstream Evaluation Protocol}
\label{ssec:tts_eval}

In each language, every system synthesizes a fixed, seeded sample of 500
held-out utterances with identical inference settings. We transcribe the audio
with a \mbox{wav2vec\,2.0} model fine-tuned for that language
\citep{xlsr2022,xlsr53} and measure character error rate (CER) against the
front-end-normalized transcript, so that no system is penalized for its own text
normalization. We also transcribe the \emph{reference} corpus audio, because a
system's CER can only be interpreted relative to the recognizer's error on
natural speech. This reference topline is 3.42\% in Bangla and 1.85\% in English.
For English, synthesis uses exactly the phoneme strings that the arms were trained
on, so the evaluation front end is identical to the training front end. We compare systems on the same test items, with paired
bootstrap confidence intervals and paired significance tests.

\begin{table*}[t]
\centering
\caption{Held-out bigram coverage (\%): the share of the bigram types occurring in
held-out utterances, which no selector could see, that each selection covers.}
\label{tab:heldout}
\small
\setlength{\tabcolsep}{4pt}
\begin{tabular}{lcccc}
\hline
& \multicolumn{2}{c}{Bangla} & \multicolumn{2}{c}{English} \\
\cmidrule(lr){2-3}\cmidrule(lr){4-5}
method & 10\% & 20\% & 10\% & 20\% \\
\hline
\textbf{Community Representative} (ours) & \textbf{99.1\,$\pm$\,0.2} & \textbf{99.4\,$\pm$\,0.1} & \textbf{98.0\,$\pm$\,0.3} & \textbf{98.8\,$\pm$\,0.2} \\
Phoneme Balance & 97.3 & 98.5 & 94.8 & 97.7 \\
Random & 94.4\,$\pm$\,0.9 & 97.3\,$\pm$\,0.5 & 94.1\,$\pm$\,0.7 & 97.1\,$\pm$\,0.3 \\
\hline
\end{tabular}

\end{table*}

\section{Results}
\label{sec:results}

\subsection{Rare-Phonotactic Coverage}
\label{ssec:coverage}

Table~\ref{tab:intrinsic} and Figure~\ref{fig:rare} show that \commrep{} leads at
every budget in both languages. At the smallest budget (10\%), it covers 89.5\%
of rare bigrams in Bangla, against 73.2\% for \phonebal{} and 49.9\% for Random,
and 73.8\% in English, against 55.0\% and 46.3\%. Its margin over \phonebal{}
narrows as the budget grows, from $+16.3$ to $+13.5$ to $+4.7$ points in Bangla
and from $+18.8$ to $+14.1$ to $+9.7$ points in English. In both languages, the
95\% confidence intervals separate at every budget.

Absolute coverage is lower in English because the budget is smaller, not because
the methods behave differently. The English pool contains 21.5 hours of audio
against 43.9 hours in Bangla, so a 10\% budget buys 2.15 hours against 4.39.
Reaching the tail of a distribution requires a certain amount of material, and
half as much audio reaches less of it. What transfers across languages is the
ordering of the methods and the size of the gaps between them.

\subsection{Coverage of the Utterance Graph}
\label{ssec:graphcov}

Figure~\ref{fig:graphcov} measures selection on the graph itself. At every budget
and in both languages, fewer utterances lie outside the one-hop reach of the
\commrep{} core-set than outside the reach of either baseline. At a 10\% budget,
its selections are within one hop of 86.1\% of Bangla utterances and 88.7\% of
English utterances, against 69.7\% and 75.7\% for \phonebal{} and 61.5\% and
60.2\% for Random. The gap appears early: at 2\%, coverage is 42.2\% against
28.0\% and 23.2\% in Bangla, and 47.7\% against 35.6\% and 22.6\% in English. At
40\%, \commrep{} covers 99.8\% and 99.9\% of the graph, while Random still misses
7.8\% and 8.8\%. Part of this lead is expected by construction, since \commrep{}
selects on this graph and the baselines do not, but it shows that its selections
reach the whole utterance graph quickly.

\paragraph{Budget across communities.} By construction, \commrep{} gives each
community of $G_u$ a share of the budget proportional to its duration, and the
baselines have no such guarantee. With seed 13 at a 10\% budget in Bangla, every
community receives between 0.98 and 1.05 times its proportional share under
\commrep{}, whereas \phonebal{} gives 3 of the 29 communities less than half of
their share and Random leaves one community unsampled.

\subsection{Held-Out Coverage and Multi-Speaker Transfer}
\label{ssec:heldout}

\paragraph{Held-out coverage.} \commrep{} also leads on held-out bigram coverage
(Table~\ref{tab:heldout}), in both languages and at both budgets, so its
advantage is not confined to the pool statistics it was built from.

\begin{table}[t]
\centering
\caption{Rare-bigram coverage (\%) on OpenSLR SLR37 (1{,}891 utterances, 6
speakers). \commrep{} leads at both budgets, and the ordering of the three
methods matches both single-speaker corpora.}
\label{tab:slr37}
\small
\setlength{\tabcolsep}{4pt}
\begin{tabular}{lcc}
\hline
method & 10\% & 20\% \\
\hline
\textbf{Community Representative} (ours) & \textbf{44.7} & \textbf{66.0} \\
Phoneme Balance & 37.3 & 54.7 \\
Random & 22.7 & 43.3 \\
\hline
\end{tabular}

\end{table}

\begin{table*}[t]
\centering
\caption{Held-out intelligibility of the TTS models trained on each arm:
character error rate (CER, \%) as mean $\pm$ 95\% CI, median CER, and the number
of catastrophic failures (items with CER${>}0.3$). Bangla arms are matched on
epochs and scored on 496 of the 500 test items; English arms are matched on
gradient steps (30{,}000 each) and scored on all 500 items. Bold marks the best trained
system per language.}
\label{tab:tts}
\small
\setlength{\tabcolsep}{4pt}
\begin{tabular}{lcccccc}
\hline
 & \multicolumn{3}{c}{Bangla} & \multicolumn{3}{c}{English} \\
\cmidrule(lr){2-4}\cmidrule(lr){5-7}
System & CER (\%) & Median & \#${>}0.3$ & CER (\%) & Median & \#${>}0.3$ \\
\hline
Reference (topline) & 3.42 $\pm$ 0.44 & 2.63 & 0 & 1.85 $\pm$ 0.21 & 0.83 & 1 \\
Random (20\%) & 4.90 $\pm$ 0.52 & 3.61 & 6 & 3.39 $\pm$ 0.38 & 1.57 & 1 \\
Phoneme Balance (20\%) & 4.62 $\pm$ 0.63 & 3.28 & 6 & 3.30 $\pm$ 0.46 & 1.35 & 1 \\
Community Representative (20\%) & \textbf{3.93 $\pm$ 0.51} & \textbf{3.06} & \textbf{0} & 2.96 $\pm$ 0.33 & 1.24 & 1 \\
Full data (100\%) & 4.47 $\pm$ 0.40 & 3.19 & 7 & \textbf{2.84 $\pm$ 0.41} & \textbf{1.11} & 3 \\
\hline
\end{tabular}

\end{table*}

\paragraph{Multi-speaker transfer.} On SLR37 (Table~\ref{tab:slr37}), \commrep{}
again outperforms both baselines by a wide margin. At a 10\% budget it covers
44.7\% of rare bigrams, against 37.3\% for \phonebal{} and 22.7\% for Random; at
20\% it covers 66.0\%, against 54.7\% and 43.3\%. Its lead over \phonebal{},
$+7.4$ points at 10\% and $+11.3$ at 20\%, falls within the $+4.7$ to $+18.8$
range observed on the two single-speaker corpora, and the ordering of the three
methods is the same. The coverage advantage therefore does not depend on the
single-speaker setting. Absolute coverage is low for
every method because the corpus is small: a 10\% budget of 2.94 hours buys less
than 18 minutes of audio.

\subsection{Downstream Intelligibility: Bangla}
\label{ssec:tts_bn}

We score 496 of the 500 test items; the remaining 4 are excluded for every system
and for the reference.

\paragraph{Comparison with the full corpus.} The 20\% core-set model reaches
3.93\% CER, against 4.47\% for the model trained on the full corpus
(Table~\ref{tab:tts}). The difference is 0.54 points (95\% CI $[+0.19,+0.84]$,
$p=0.011$), and the median CER is also lower (3.06\% against 3.19\%). The
core-set model achieves this with $4.5\times$ less training time.

\paragraph{Comparison with Random.} At the same budget, the core-set model
improves on the random subset by 0.97 CER points (95\% CI $[+0.61,+1.35]$,
$p<0.001$), the largest difference between any two of the 20\% arms. Random
reaches 4.90\% CER, above the full corpus (4.47\%), while the core-set (3.93\%) is
the closest of all trained systems to the 3.42\% reference topline. Which 20\% of
the data is selected therefore matters more than training on five times as much
audio. The systems also differ in the error tail. Like the reference audio, the
core-set model produces \emph{no} utterance with CER above 0.3, and its worst case
is 29\%. \phonebal{} produces 6 such failures, Random produces 6 with a worst case
of 65\%, and the full corpus produces 7 with a worst case of 67\%. Word error rate
shows the same ordering at lower resolution: 21.06\% for the core-set, 21.7\% for
the full corpus, 22.4\% for \phonebal{}, and 23.3\% for Random, against a 19.5\%
topline.

\paragraph{Comparison with \phonebal{}.} This arm tests whether graph structure
helps beyond an explicit phonetic-balance criterion, not only beyond chance.
\phonebal{} falls between the core-set and Random: its CER of 4.62\% is lower than that of Random (4.90\%) but
higher than those of the core-set and the full corpus. Its median CER is 3.28\%,
against 3.06\% for the core-set. The core-set advantage is $+0.69$ CER points
(95\% CI $[+0.36,+1.02]$, Wilcoxon $p=0.007$) and $+1.34$ WER points ($p=0.014$),
by paired bootstrap over the same 496 items. \commrep{} therefore outperforms
entropy-based selection at an identical duration budget. This is the comparison that the intrinsic coverage results
predict, and it separates the selection criterion from the budget.

\subsection{Downstream Intelligibility: English}
\label{ssec:tts_en}

No English test item has a corrupt reference, so all 500 items are scored. The
core-set model leads both same-budget controls on the mean and on the median
(Table~\ref{tab:tts}): 2.96\% CER against 3.30\% for \phonebal{} and 3.39\% for
Random, and 1.24\% against 1.35\% and 1.57\% on the median. The three selectors
are ordered as in Bangla and as the intrinsic coverage results predict, and both
differences are significant: $+0.43$ CER points over Random (95\% CI
$[+0.20,+0.68]$, $p=0.006$) and $+0.34$ points over \phonebal{} (95\% CI
$[+0.11,+0.54]$, Wilcoxon $p=0.018$). The intrinsic advantage therefore converts
into a measurable gain in intelligibility in both languages. The English margins
are about half the size of the Bangla ones (\S\ref{ssec:headroom}).

The full pool, trained for the same number of updates as the subsets, reaches the
lowest English CER: 2.84\%, against 2.96\%, 3.30\%, and 3.39\%. It clearly
outperforms Random ($+0.55$ points, $p=6\times10^{-6}$). This is the expected
outcome when all arms receive the same number of updates (\S\ref{ssec:epochs}).

\paragraph{Inference speed.} All systems in both languages run much faster than
real time on an Apple M4 Pro (median real-time factor between 0.040 and 0.053),
and their output durations closely match the references (mean duration ratio
1.00 to 1.02 in Bangla and 0.95 to 0.99 in English).

\section{Discussion}
\label{sec:discussion}

\subsection{Why the English Margins Are Smaller}
\label{ssec:headroom}

The English gains over Random and \phonebal{} (0.43 and 0.34 points) are about
half the Bangla gains (0.97 and 0.69 points). The most likely reason is limited
headroom. Every English system lies between 2.8\% and 3.4\% CER against a 1.85\%
topline, so only about 1.5 points separate the worst system from the ceiling,
while the 95\% confidence intervals of the individual systems on 500 items have
half-widths between 0.21 and 0.46 points. Much of the
Bangla margin also comes from the error tail, which English almost lacks: one
item above 30\% CER for the reference and for each 20\% arm, and three for the
full-pool arm, against 6 for the Bangla random arm and 7 for the Bangla full
corpus. When almost every item is already intelligible, a selection method has
fewer failures left to prevent, so its margins shrink, although they remain
significant.

\subsection{Equal Epochs Versus Equal Updates}
\label{ssec:epochs}

The two training protocols answer related but different questions. In Bangla,
the arms are matched on epochs, so the full-corpus model receives about four
times as many gradient steps as the core-set model (454{,}400 against 109{,}400).
Under this protocol, the core-set model is both $4.5\times$ cheaper to train and
more intelligible than the full-corpus model (\S\ref{ssec:tts_bn}). In English,
all arms receive the same 30{,}000 updates, and the full pool keeps a small lead
over the core-set (2.84\% against 2.96\% CER), as expected when a larger and more
varied training set is trained for the same number of steps. We therefore do not
claim that a core-set outperforms all of the data at equal compute. Our claim is
narrower. At any fixed duration budget, \commrep{} outperforms
both random and entropy-based selection, and when training runs a fixed number
of passes over its data, a well-chosen 20\% subset can exceed the quality of the
full corpus at a fraction of the cost, as it does in Bangla.

\subsection{What Replicates Across Languages}
\label{ssec:replicates}

The intrinsic results replicate fully: the graph structure is present in both
corpora, and \commrep{} improves coverage over both baselines in both. The downstream results also
replicate. In both languages, the three same-budget selectors are ordered in the
same way (\commrep{}, then \phonebal{}, then Random), and the core-set model is
significantly more intelligible than both baselines. Two aspects differ between
the languages. First, the margins are about half as large in English, where every
system is close to the recognizer's floor (\S\ref{ssec:headroom}). Second, the
comparison with the full data depends on the training protocol: the core-set
outperforms the full corpus when the arms are matched on epochs (Bangla), but not
when they are matched on updates (English; \S\ref{ssec:epochs}).

\section{Conclusion}
\label{sec:concl}

We showed that a TTS corpus, represented as a phonotactic graph of utterances,
has measurable network structure in two typologically distant languages, and
that this structure survives degree-preserving randomization. A selector that
samples within graph communities covers more rare phonotactic patterns than
random and entropy-based selection at every budget in both languages, on held-out
data, and on a third, multi-speaker corpus, and it reaches the whole utterance
graph faster than either baseline. In
downstream TTS training at a 20\% budget, its core-sets produce significantly
more intelligible models than random or entropy-based selection in both Bangla
and English. In Bangla, the core-set model also outperforms full-corpus training
by 0.54 CER points with $4.5\times$ less training time. A natural next step is to
repeat the downstream comparison on a multi-speaker corpus with a wider range of
recording quality, where better coverage has more to offer.

\section*{Limitations}

Our claims have several limitations. Both main corpora are \textbf{single
speaker}, each read by one person, so speaker and prosodic diversity are absent
by construction, and no objective includes a speaker-coverage term.
\S\ref{ssec:heldout} shows that the intrinsic results transfer to real
multi-speaker data, but this check is intrinsic only; we did not repeat the
downstream comparison there. The G2P is rule-based for Bangla and
dictionary-based for English. Its errors affect every method equally, which keeps
the comparison fair, but absolute coverage numbers inherit its accuracy. The two
downstream protocols differ: the Bangla arms are matched on epochs and the English
arms on gradient steps, so the two languages answer related but different
questions, and the advantage over the full corpus holds only under the Bangla
protocol. One training hyperparameter also differs across the Bangla arms
(\S\ref{ssec:tts_train}). The English margins are about half the size of the
Bangla ones, and with 500 test items against a 1.85\% recognizer floor, their
confidence intervals are wide relative to the effect sizes. Finally, the heavy
degree tail of $G_u$ (Figure~\ref{fig:degree}) is partly \emph{hubness}, a known
artifact of $k$-NN graphs in high dimensions, and we make no power-law claim.

\section*{Ethics Statement}

The Bangla sentences come from Common Voice Scripted Speech 26.0 -- Bengali,
released under CC0-1.0, and were recorded by an in-house speaker who gave consent
for their use in this research. The recordings are in-house and not publicly
released, and no synthetic speech was used in any corpus. LJSpeech is in the
public domain, and OpenSLR SLR37 is distributed under the CC BY-SA 4.0 licence.
Core-set selection lowers the data and compute needed to build a TTS voice. It
therefore also lowers the cost of building a voice without the speaker's consent,
so core-sets should be drawn only from corpora whose speakers permit synthesis
use.

\bibliography{refs}

\end{document}